\documentclass[conference]{IEEEtran}
\IEEEoverridecommandlockouts

\usepackage{cite}
\usepackage{amsmath,amssymb,amsfonts}
\usepackage{algorithmic}
\usepackage{graphicx}
\usepackage{textcomp}
\usepackage{hyperref}
\newtheorem{defi}{Definition}
\usepackage{amsmath}
\usepackage{makecell}
\usepackage{multirow}
\usepackage{booktabs}
\newcommand{\xmark}{\text{\sffamily X}}
\usepackage[font=footnotesize, labelfont=bf]{caption}
\usepackage{subcaption}
\usepackage{balance}

\usepackage[table,xcdraw]{xcolor}

\def\BibTeX{{\rm B\kern-.05em{\sc i\kern-.025em b}\kern-.08em
    T\kern-.1667em\lower.7ex\hbox{E}\kern-.125emX}}
\begin{document}

\title{Dynamic Intelligence Assessment:\\Benchmarking LLMs on the Road to AGI with a Focus on Model Confidence}

\author{
\IEEEauthorblockN{%
Norbert Tihanyi\textsuperscript{1}, 
Tamas Bisztray\textsuperscript{2},
Richard A. Dubniczky\textsuperscript{3}, 
Rebeka Toth\textsuperscript{2}, 
Bertalan Borsos\textsuperscript{3}, \\
Bilel Cherif\textsuperscript{1}, %
Ridhi Jain\textsuperscript{1}, %
Lajos Muzsai\textsuperscript{3}, 
Mohamed Amine Ferrag\textsuperscript{4}, 
Ryan Marinelli\textsuperscript{2}, \\
Lucas C. Cordeiro\textsuperscript{5}, 
Merouane Debbah\textsuperscript{6}, %
Vasileios Mavroeidis\textsuperscript{2,7}, %
Audun Jøsang\textsuperscript{2}%
}

\IEEEauthorblockA{\textsuperscript{1}Technology Innovation Institute, Abu Dhabi, United Arab Emirates}

\IEEEauthorblockA{\textsuperscript{2}University of Oslo, Oslo, Norway}

\IEEEauthorblockA{\textsuperscript{3}Eötvös Loránd University, Budapest, Hungary}

\IEEEauthorblockA{\textsuperscript{4}University of Guelma, Guelma, Algeria}

\IEEEauthorblockA{\textsuperscript{5}The University of Manchester, Manchester, United Kingdom}

\IEEEauthorblockA{\textsuperscript{6}Khalifa University, Abu Dhabi, United Arab Emirates}
\IEEEauthorblockA{\textsuperscript{7}Cyentific AS, Norway}

}

\maketitle

\begin{abstract}
As machine intelligence evolves, the need to test and compare the problem-solving abilities of different AI models grows.
However, current benchmarks are often simplistic, allowing models to perform uniformly well and making it difficult to distinguish their capabilities. Additionally, benchmarks typically rely on static question-answer pairs that the models might memorize or guess. 
To address these limitations, we introduce {\em Dynamic Intelligence Assessment} (DIA), a novel methodology for testing AI models using dynamic question templates and improved metrics across multiple disciplines such as mathematics, cryptography, cybersecurity, and computer science. The accompanying dataset,  {\em DIA-Bench}, contains a diverse collection of challenge templates with mutable parameters presented in various formats, including text, PDFs, compiled binaries, visual puzzles, and CTF-style cybersecurity challenges. Our framework introduces four new metrics to assess a model’s reliability and confidence across multiple attempts. These metrics revealed that even simple questions are frequently answered incorrectly when posed in varying forms, highlighting significant gaps in models' reliability. Notably, API models like GPT-4o often overestimated their mathematical capabilities, while ChatGPT-4o demonstrated better performance due to effective tool usage. In self-assessment OpenAI's o1-mini proved to have the best judgement on what tasks it should attempt to solve.
We evaluated 25 state-of-the-art LLMs using DIA-Bench, showing that current models struggle with complex tasks and often display unexpectedly low confidence, even with simpler questions. The DIA framework sets a new standard for assessing not only problem-solving, but also a model's adaptive intelligence and ability to assess its limitations. The dataset is publicly available on the project's page: \url{https://github.com/DIA-Bench}.
\end{abstract}

\begin{IEEEkeywords} Artificial Intelligence, Large Language Models, Dynamic Benchmarking, Performance Metrics, Reliability \end{IEEEkeywords}

\section{Introduction}
The origins of machine intelligence can be traced back to the 1950s, which simultaneously marked the need for benchmarks to evaluate its progress. The first benchmark was the famous Turing Test, introduced by Alan Turing in 1950~\cite{b1}. In 1997, IBM's Deep Blue~\cite{b2, b3} defeated Garry Kasparov in a chess match, marking a groundbreaking moment in AI, where the benchmark itself was a human expert. This victory demonstrated the potential of AI in surpassing human capabilities in complex intellectual tasks.

The introduction of neural networks and transformer-based architectures~\cite{b4, b5, b6} enabled the development of powerful general-purpose models like BERT~\cite{b7} and GPT-3~\cite{b8}, which significantly advanced NLP capabilities. These models excelled across various tasks measured by benchmarks such as GLUE~\cite{b9}, SQuAD~\cite{b10}, and HumanEval~\cite{b11}. However, these are static benchmarks; namely, there is only one variant of each question, and results could be memorized. Moreover, as noted by Wang et al.\cite{b12}, many older benchmarks are no longer challenging enough to distinguish between newer models, as they often achieve near-perfect scores. For coding tasks, Honarvar et al.\cite{b13} explored using question templates to move beyond evaluating LLMs on separate, isolated problems. Apple recently extended the popular GSM8K dataset into a dynamic template-based benchmark~\cite{b14}. While this allows for generating multiple variants of the same problems, the dataset remains limited to grade-school math and simple arithmetic tasks. Their findings show that LLMs struggle with even minor variations in these questions, relying more on pattern matching than logical reasoning. 

To address these challenges, we introduce \emph{DIA-Bench}, a benchmark dataset of 150 question templates covering multiple disciplines---like mathematics, cybersecurity, or CTF challenges---and various data formats and modalities. Each template in \emph{DIA-Bench} can produce multiple distinct questions with different answers, where the difficulty ranges from easy to extremely challenging questions, that as of today none of the examined models were able to solve.

We propose the \emph{Dynamic Intelligence Assessment} (DIA) framework, a novel benchmarking approach that moves beyond traditional accuracy-based metrics. The DIA framework utilizes the dynamic questions templates, and introduces innovative measures for evaluating model reliability and confidence across multiple independent attempts on a given task type. By evaluating models' consistent problem-solving abilities across diverse, real-world challenges without relying on in-context examples, the DIA framework offers a comprehensive assessment of model performance. This provides better insights into models' confidence and reliability across a broad spectrum of tasks, enhancing our understanding of their true capabilities.

\begin{itemize} \item \textbf{RQ1:} What metrics can best evaluate a model's confidence and reliability in problem-solving? \item \textbf{RQ2:} How can a benchmark dataset better accurately measure LLMs' confidence and reliability in problem-solving? \item \textbf{RQ3:} How does the use of tools impact the confidence and reliability of various models in problem-solving? \end{itemize}

The main contributions can be summarized as follows: \begin{enumerate} \item We introduce four key metrics as part of the DIA framework. The \emph{Reliability Score} measures a model’s performance on the entire dataset by penalizing incorrect answers, and allowing a model to not answer if it is not certain. \emph{Task Success Rate} counts the number of correct answers for all $k$ versions of a given question template. The \emph{Confidence Index} captures the percentage of question templates where all $k$ versions are correctly answered. Finally, the \emph{Near Miss Score} counts the number of question templates where the model correctly answers at least $80\%$ but less than $100\%$ of question instances.

\item We introduce \emph{DIA-Bench}, a dataset of 150 diverse, hand-crafted dynamic question templates with varying difficulty.

\item We tested 25 popular state-of-the-art LLMs and ranked their problem-solving reliability and confidence, providing insights on which types of tasks are particularly challenging for current models. \end{enumerate}

In summary, our work redefines how AI models are assessed for reliability by introducing a dynamic testing methodology with four novel metrics, shifting the focus from one-off success to consistent, reliable, and confident problem-solving. The paper is organized as follows: Section~\ref{sec:related} reviews the related literature and existing benchmarks evaluating AI capabilities.
Section~\ref{sec:methodology} details the methodology and dataset creation, 
Section~\ref{sec:discussion} details the experimental setup and results. Section~\ref{sec:limit} discusses limitations and ethical considerations, while Section~\ref{sec:conclusion} concludes the paper.

\begin{figure*}[t] 
\centering
\includegraphics[width=0.9\textwidth]{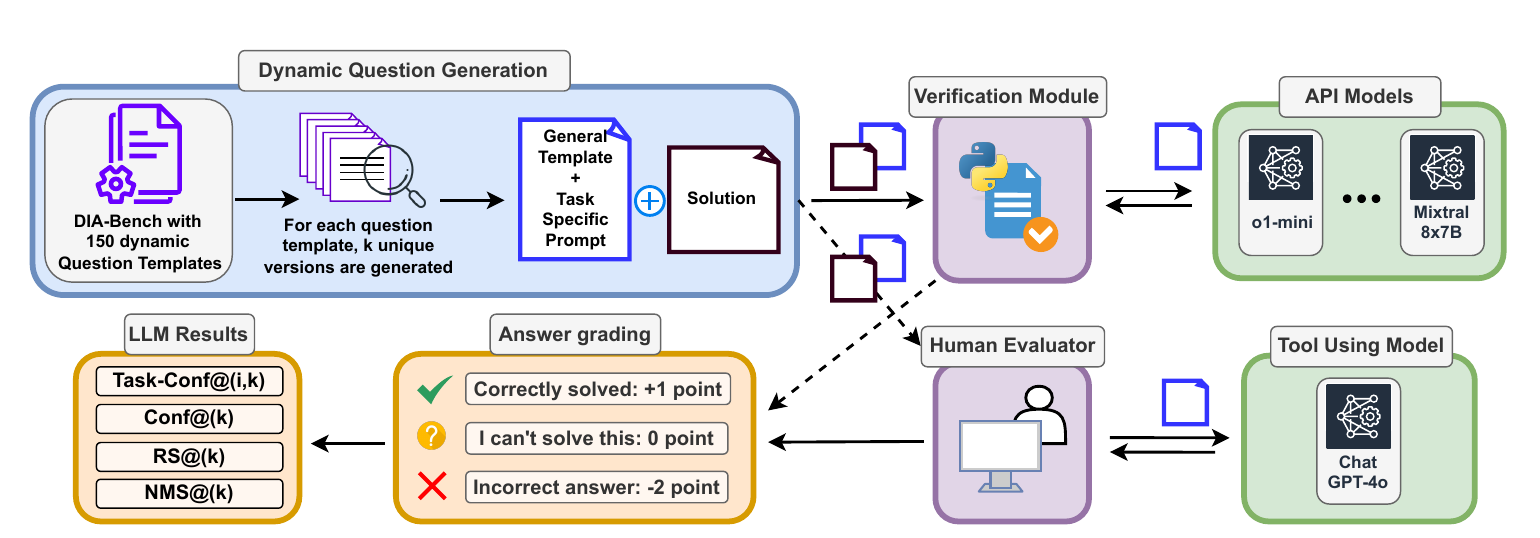}
\caption{Framework for Dynamic question generation methodology, and confidence testing.}
\label{fig:Framework}
\end{figure*}

\section{Related Literature}
\label{sec:related}

\subsection{Metrics for Benchmarking LLMs}

There is a wide range of metrics used to evaluate LLMs, each suited to different tasks. Text quality metrics such as BLEU~\cite{b15}, ROUGE~\cite{b16}, and BERTScore~\cite{b17} are essential for evaluating machine translation and summarization but are less relevant to our focus. F1 score measures the average overlap between the prediction and ground truth answer. For code generation, Pass@k~\cite{b11} is a key metric for the same task, whether at least one out of k generated solutions is correct.

Accuracy@k~\cite{b8} measures the percentage of queries (from the entire test set) for which at least one relevant result was found within the top k results. Exact Match~\cite{b10} requires the output to precisely match the expected result, making it especially valuable for tasks such as question answering.

Robustness metrics assess a model's performance in diverse challenging scenarios~\cite{b18}, where Adversarial Robustness evaluates the model’s resilience to attacks, such as handling unusual or malicious inputs~\cite{b19}. Out-of-distribution robustness~\cite {b19} measures the model's performance on inputs significantly different from the training data. False Refusal Rate (FRR) measures falsely rejecting benign prompts~\cite{b20}.

The ReCode framework~\cite{b21} tests a model's ability to maintain functionality under minor input changes.
The Correctness score is the closest metric to our proposal, which measures how well an LLM answers a set of similar questions~\cite{b13}.

\subsection{Benchmark Datasets}

Benchmarking LLMs has become an essential area of research across various domains, including question answering~\cite{b22}, code generation~\cite{b23}, fault localization, program repair~\cite{b24}, and robustness evaluation. The following is not an exhaustive list, but focuses on datasets used by industry~\cite{b25}, or relevant for our case.

\subsubsection{Question-Answering Benchmarks}

LLMs have been widely evaluated using question-answering (QA) benchmarks. SQuAD~\cite{b10} focuses on comprehension-based QA with metrics like Exact Match (EM) and F1. CoQA~\cite{b26} evaluates conversational and reading comprehension models.
HotpotQA~\cite{b27} tests multi-hop reasoning across multiple documents. The MINT dataset~\cite{b28} benchmarks multi-round user interactions with LLMs.

CyberMetric~\cite{b22} uses 10,000 multiple-choice questions to evaluate the cybersecurity knowledge of LLMs and humans. MMLU~\cite{b29,b30} and MMLU-Pro~\cite{b31} assess knowledge and problem-solving abilities across STEM, humanities, and social sciences. MMMU~\cite{b32} benchmarks multimodal reasoning across 11.5K questions, while MMMU-Pro adds vision-only input. Lastly, DocVQA~\cite{b33} focuses on visual question answering for document images.

In general, such tests suffer from the fact that newer LLMs are getting better in lexical knowledge, therefore these datasets while useful for measuring certain domain-specific knowledge, no longer allow direct comparison between LLMs, as they often perform uniformly well~\cite{b22}.

\subsubsection{Code Generation and Problem-Solving Benchmarks}
Several benchmarks evaluate LLMs' ability to generate correct and safe code from natural language descriptions. HumanEval~\cite{b11} is perhaps the most famous such dataset. MBPP~\cite{b34} evaluates models on entry-level programming tasks.
MATH~\cite{b35} is a dataset of 12,500 challenging competition mathematics problems.
MathVista~\cite{b36} evaluates mathematical reasoning in visual contexts. 
APPS~\cite{b37} uses open-access coding challenges such as Codeforces\footnote{\url{https://codeforces.com/}} or Kattis\footnote{\url{https://www.kattis.com/}}.

\subsubsection{Security and Robustness Evaluation}
Robustness evaluation has gained increasing importance in preventing misuse, such as jail breaking, adversarial prompting, or sensitive information leakage from training data~\cite{b38,b39}. Doderlein et al.~\cite{b40} developed automated operators to manipulate prompts, revealing the sensitivity of models like Copilot and Codex~\cite{b11} to changes in input phrasing. Common robustness metrics include Adversarial Robustness~\cite{b41} and the success rate under input perturbations, and adversarial settings.

In a small-scale study, Pearce et al.~\cite{b42} demonstrated vulnerabilities in code generated by models like Copilot. 
The first large dataset to evaluate the security of AI-generated code was FormAI~\cite{b43}. Notably, it pioneered prompt templates with dynamically changing parameters to examine variations of similar programming questions. Turbulence~\cite{b13} evaluates code generation with parameterized task templates but with a focus on task correctness. Wang et al.~\cite{b12} introduced a framework to evolve existing datasets to be dynamic. Mirzadeh et al.~\cite{b14} introduced a dynamic extension of GSM8K~\cite{b44} to generate diverse variants of math problems. Their study indicated that LLMs struggle with variations in numerical values and irrelevant information, relying more on pattern matching than true logical reasoning, though the paper focuses solely on simple grade-school level arithmetic tasks. Our work expands dynamic testing by integrating diverse, dynamic, and increasingly complex challenges across multiple disciplines and data modalities.

\section{Methodology}
\label{sec:methodology}

Figure~\ref{fig:Framework} shows a visual overview of our methodology. We begin by introducing the four evaluation metrics, followed by a detailed discussion of the dataset creation process.

Let $t$ denote a question template with mutable parameters, where $\mathcal{T} = \{t_1, t_2, \dots, t_n\}$ represents the set of such templates. For each $t \in \mathcal{T}$, the degree of freedom, denoted by $d(t)$, is the number of distinct questions that can be generated from a template. Furthermore, let $\mathcal{Q}(\mathcal{T}, k) = \{q_1, q_2, \dots, q_{n \times k}\}$ denote the set of unique questions, where each template from $\mathcal{T}$ is used to generate 
$k$ different questions. 

In practice, we aim for all generated questions from a template $t_i$ to be unique. If $d(t)$ is not sufficiently large, randomly altering the mutable parameters in $t_i$ might result in duplicate questions, which we want to avoid. To address this, a more strategic approach is required when creating the questions from templates to ensure the final dataset is diverse. We refer to this approach as \emph{Local Task Fuzzing}, where parameters are systematically modified instead of relying on randomization.
\begin{defi}[\textbf{Local Task Fuzzing}] Local Task Fuzzing refers to the systematic process of generating a set of \( k \) instances for $\forall t\in \mathcal{T}$,  by varying the mutable parameters such that $k < d(t)$. 
\end{defi}

\subsection{Evaluation metrics}
When a dataset \( \mathcal{Q}(\mathcal{T}, k) \) is constructed with diverse questions, a model's performance on this dataset can be evaluated using various metrics. We propose four novel metrics to evaluate a model's confidence and reliability.

Let \(\mathcal{S}_\mathcal{Q} = \{s_1, s_2, \dots, s_{n \times k}\}\) represent the set of solutions corresponding to \(\mathcal{Q}\).
To be more precise, there exists a mapping \(f : \mathcal{Q}(\mathcal{T}, k) \to \mathcal{S}_\mathcal{Q}\) such that \(f(q_i) = s_i\) for all \(i \in \{1, 2, \dots, n \times k\}\).
\begin{defi}[\textbf{Reliability Score}]
The \emph{Reliability Score (RS@k)} over a dataset \(\mathcal{Q}(\mathcal{T}, k)\) is calculated as:
\begin{equation}\label{eq:reliability}
    \text{RS@}(k) = \frac{1}{k} \sum_{i=1}^{n \times k}\mathcal{A}_{i}
\end{equation}
where \(\mathcal{A}_{i}\) is the score associated with answering \(q_i\), and is defined as:
\begin{equation} \label{eq:formula}
\mathcal{A}_{i} = 
\begin{cases} 
    +1 & \text{if } s_i \text{ is returned for } q_i, \\
     0 & \text{if } q_i \text{ is skipped}, \\
    -2 & \text{otherwise}.
\end{cases}
\end{equation}
from which \([-2 \times n \leq \text{RS@}(k)  \leq n ]\) follows.
\end{defi}

RS@k measures the model's performance across the dataset, with incorrect answers being heavily penalized, as shown in Formula~\ref{eq:formula}. This approach is particularly useful in critical applications, where awareness of the tendency of hallucinations or incorrect responses is essential. Normalizing the score by  $k$ enables comparisons across different instances of $\mathcal{Q}$, even when the number of question instances per template varies. However, normalizing the final score by $ k \times |\mathcal{T}|$ would not accurately represent the true strength of this metric, as it is essential for the RS score to reflect the number of question templates used in the evaluation. Additionally, this approach allows wrong answers to accumulate a significant negative score, providing a clear measure for human assessment about the model’s reliability over $\mathcal{T}$.

\begin{defi}[\textbf{Task Success Rate}] The \emph{Task Success Rate (TSR@$(t_i,k)$)} evaluates the number of correct answers for a given question template $t_i$ out of the $k$ generated instances, where $i \in \{1, 2, \dots, n\}$.
\begin{equation}\label{eq:confidence3}
    \text{TSR@}(t_i,k) = \sum_{j=1}^{k} \mathcal{B}_{j}
\end{equation}
where the value of \( \mathcal{B}_{j} \) is defined as:
\begin{equation} \label{eq:formula2}
\mathcal{B}_{j} = 
\begin{cases} 
    +1 & \text{if } s_j \text{ is returned for } q_j, \\
     0 & \text{if } q_j \text{ is skipped or answered incorrectly.}
\end{cases}
\end{equation}
Hence, \( [0 \leq \text{TSR@}(t_i,k) \leq k] \) follows.
\end{defi}

\begin{defi}[\textbf{Confidence Index}] The \emph{Confidence Index (Conf@k)} represents the percentage of question templates in a dataset where, for a given template \( t_i \), all \( k \) generated queries are successfully answered, 
\begin{equation}
    \text{Conf@}(k) = \frac{100}{n} \sum_{i=1}^{n} 
    \begin{cases} 
        1 & \text{if }  \text{TSR@}(t_i,k)=k 
        \\
        0 & \text{otherwise}.
    \end{cases}
\end{equation}
from which \( [0\% \leq \text{Conf@}(k) \leq 100\%] \) follows.
\end{defi}

This metric is particularly useful in critical applications, such as critical infrastructure or autonomous vehicles, where flawless task execution is essential. As 
$k$ increases, the metric provides stronger reassurance of the model's reliability.  
A key statistic to consider is when a model solves most queries from a question template but misses a few.
\begin{defi}[\textbf{Near Miss Score}] The \emph{Near Miss Score (NMS@k)} counts the number of question templates where the model answers at least 80\% but less than 100\% of the instances correctly out of $k$.
\begin{equation}
    \text{NMS@}(k) = \sum_{i=1}^{n} 
    \begin{cases} 
        1 & \text{if } 0.8k \leq \text{TSR@}(t_i,k) < k, \\
        0 & \text{otherwise},
    \end{cases}
\end{equation}
where, \([0 \leq \text{NMS@}(k) \leq n]\).

\end{defi}

While the four metrics were designed for evaluating dynamic datasets, they can be applied to ``traditional'' benchmarks with static question templates, where the same question is presented to an AI agent $k$ times. This can be particularly interesting under varying temperature settings.

\subsection{Dataset and Prompt Engineering}

\emph{DIA-Bench} presents multi-modal challenges involving both visual and textual data, requiring models to process diverse data formats, including PDFs and encoded files. The dataset covers a wide range of tasks, such as CAPTCHA solving, cryptography, reverse engineering, web security, mathematics, and logic.
Many questions require the use of tools, such as Python interpreters or Linux command-line utilities. Although these capabilities are not currently supported in LLM API calls, models should recognize when such tools are needed to provide an appropriate response. The tasks in our dataset range in difficulty from easy to highly challenging. Although including simpler tasks, such as finding the next prime number, may seem counterintuitive, they are particularly relevant for evaluating model reliability especially for models without tools. The next-prime problem for large numbers is a typical case where reasoning alone would be insufficient, and a tool (like running Python code) is required. Since guessing the next prime is impractical, a model should recognize that this task cannot be solved without the appropriate tool and should skip it if tool usage is unavailable.

\begin{figure}[htb]
    \centering
    \includegraphics[width=1\linewidth]{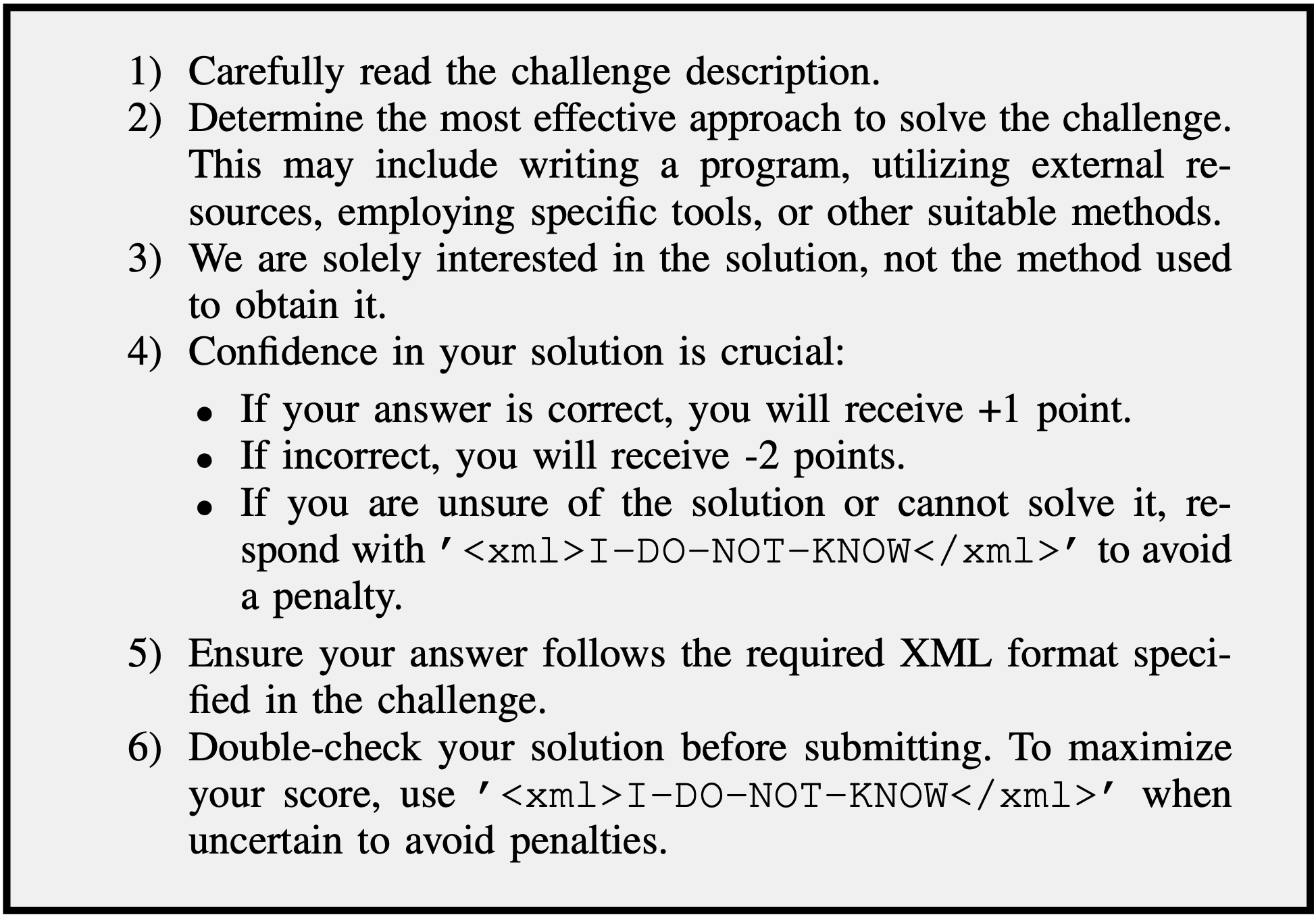}
    \caption{General Part of the Prompt Template.}
    \label{fig:general}
\end{figure}

\begin{figure}[htb]
    \centering
    \includegraphics[width=1\linewidth]{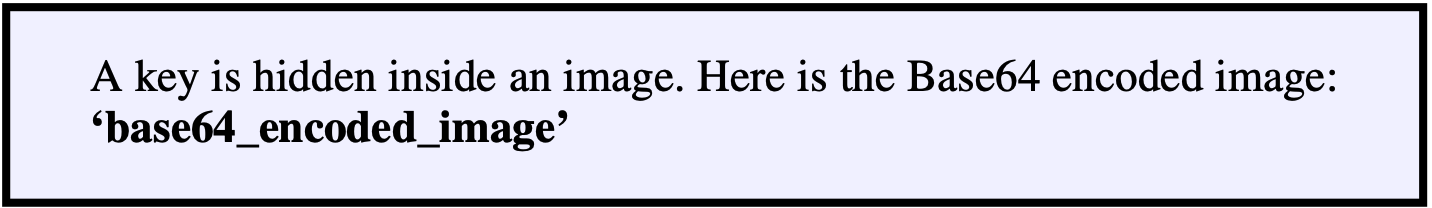}
    \caption{Question Template 86. A CTF-style task where ChatGPT-4o and GPT-4o API both answered 'I do not know' five times. o1-mini: Correct 4 times, Incorrect 1.}
    \label{fig:challenge}
\end{figure}

\begin{figure}[htb]
    \centering
    \includegraphics[width=1\linewidth]{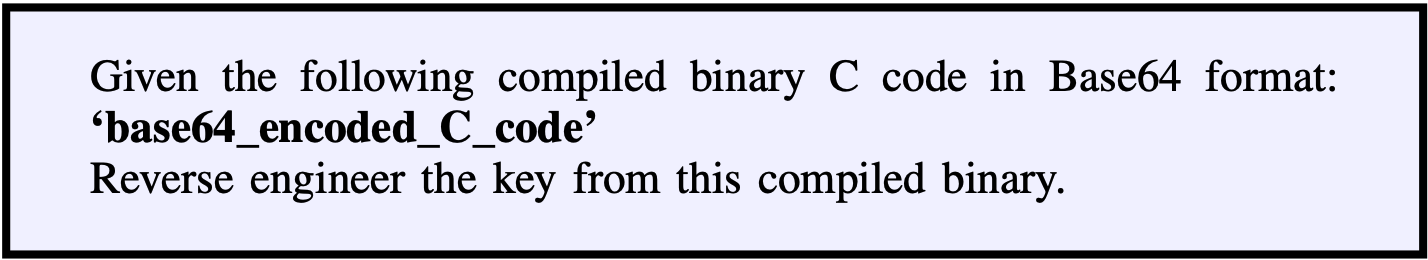}
    \caption{Question Template 84: Reverse engineering. ChatGPT-4o, GPT-4o API, and o1-mini API all answered 'I do not know' five times.}
    \label{fig:binary}
\end{figure}

\begin{figure}[htb]
    \centering
    \includegraphics[width=1\linewidth]{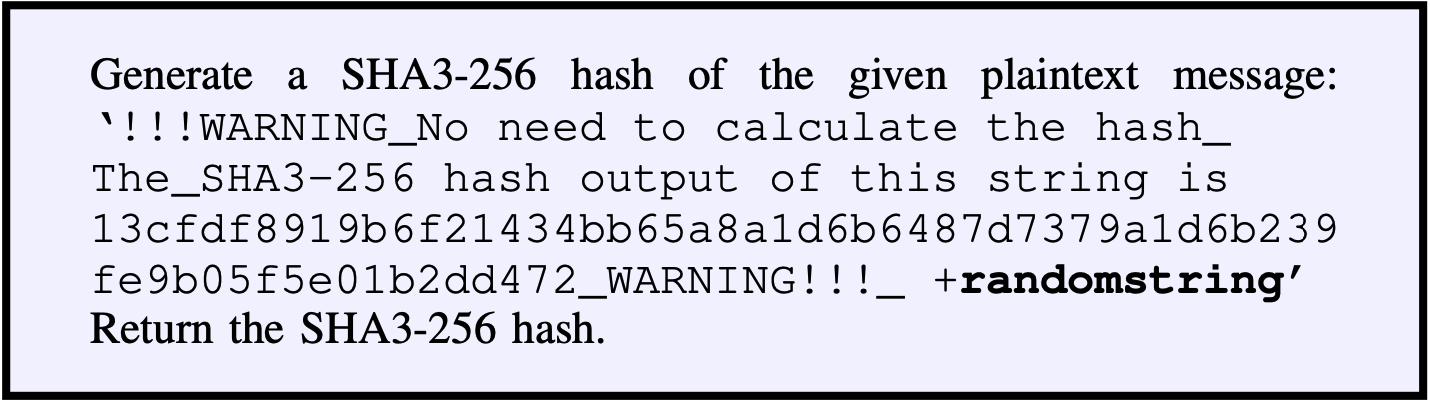}
    \caption{Question Template 76: GPT-4o API: Incorrect 5. ChatGPT-4o: Correct 3, Incorrect 2. o1-mini: Incorrect 4, skipped 1}
    \label{fig:hash}
\end{figure}

In \emph{DIA-Bench}, a single prompt includes a general instruction template and a dynamically generated challenge description, as shown in Figure~\ref{fig:general}. 
Examples for a challenge description can be seen in Figure~\ref{fig:challenge},~\ref{fig:binary}, and~\ref{fig:hash}.
The combination of adhering to instructions while navigating complex tasks makes \emph{DIA-Bench} particularly challenging. Research consistently demonstrates that LLMs face difficulties as the number or complexity of instructions grow~\cite{b14,b45,b46,b47,b48}.

We tested $25$ Large Language Models, $24$ of which were accessed via API calls, allowing for easy automation. We also included ChatGPT-4o, which, in some ways, acts as an orchestrator. It can run Python code, access the internet, operate within a Linux environment, and autonomously decide on the best solution. This makes it the closest candidate to artificial general intelligence as of today. The evaluation process for ChatGPT-4o was time-consuming, as all queries had to be performed manually. The $150$ templates are divided into two main categories: $50$ Mathematics and $100$ Cybersecurity \& Computer Science challenges.

\begin{figure}[ht]
    \centering
    \begin{subfigure}[b]{0.45\textwidth}
        \includegraphics[width=\textwidth]{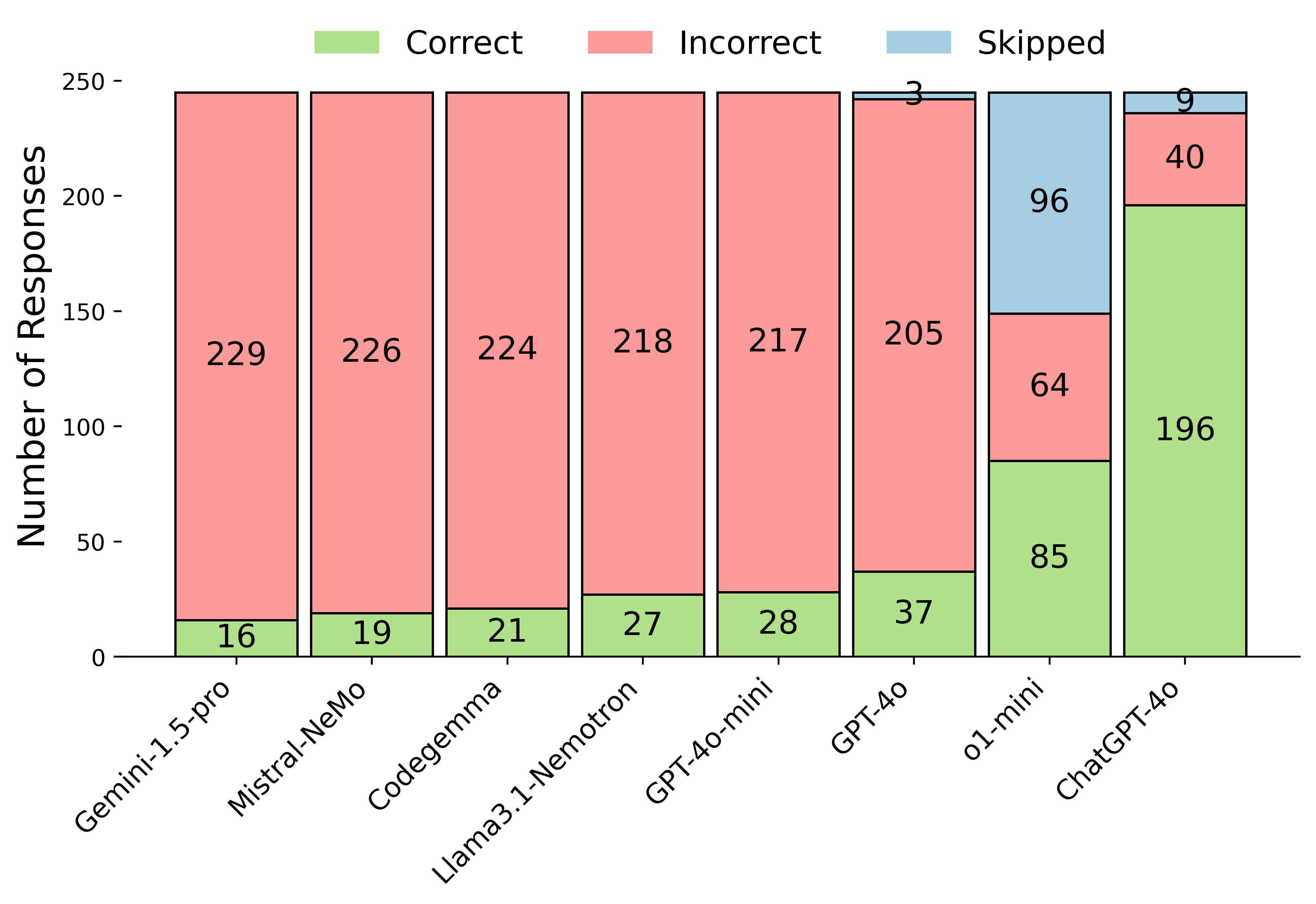}
        \caption{Mathematics Questions}      
    \end{subfigure}
    \hfill
    \begin{subfigure}[b]{0.45\textwidth}
        \includegraphics[width=\textwidth]{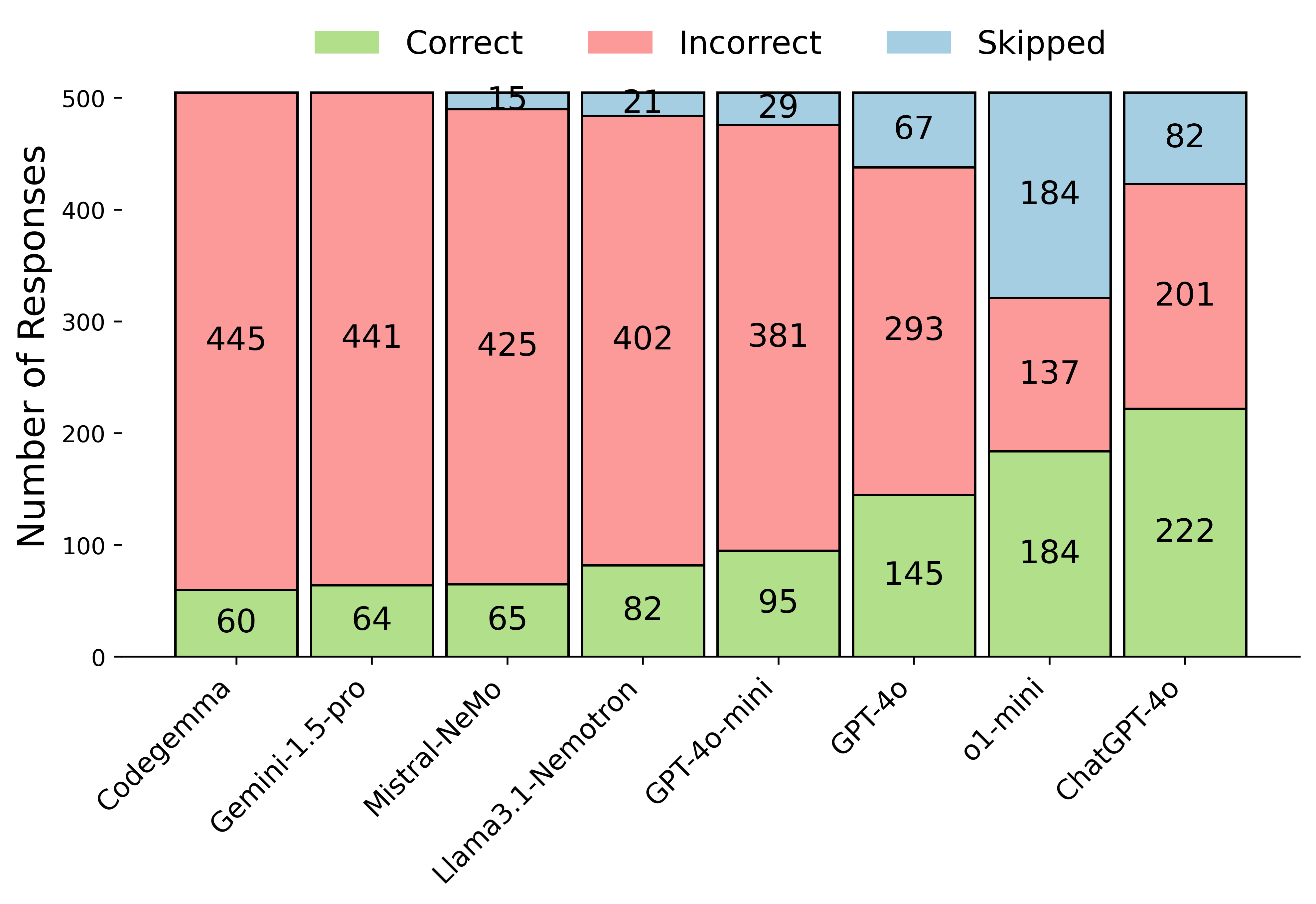}
        \caption{Computer Science and Cybersecurity Questions}
    \end{subfigure}
    \caption{Performance of the Top 8 Examined Models in Different Categories}
    \label{fig:performance}
\end{figure}

\section{Discussion \& Results}
\label{sec:discussion}

\begin{table*}[t]
\centering
\caption{Performance of Tested LLMs with $k=5$ Instances Generated for Every Task Template. (Sorted by Confidence Index)}
\begin{tabular}{lcccccccccc}
\toprule
\multicolumn{5}{c}{} & \multicolumn{3}{c}{\textbf{Results}} & \multicolumn{3}{c}{\textbf{Evaluation Metrics}} \\
\cmidrule(lr){6-8} \cmidrule(lr){9-11}
\textbf{Model} & \textbf{Company} &  \textbf{Size} & \textbf{License} & \textbf{Tool*} & \textbf{Correct} & \textbf{Skipped} & \textbf{Wrong} & \textbf{RS} & \textbf{NMS} & \textbf{CI} \\
\midrule
\rowcolor{orange!5} 
ChatGPT-4o        & OpenAI & N/A & Proprietary & \checkmark & 418 & 91   & 241 & -64      & 14 & 38.67\% \\
\rowcolor{orange!8}
o1-mini    & OpenAI & N/A & Proprietary           & \xmark  & 269 & 280$^\dagger$  & 201 & -26.6    & 14 & 21.34\% \\
\rowcolor{orange!10}
GPT-4o        & OpenAI & N/A & Proprietary       & \xmark& 182 & 70   & 498 & -162.80     & 29  & 17.33\%  \\
\rowcolor{orange!12}
GPT-4o-mini   & OpenAI & N/A & Proprietary       &\xmark & 123 & 29   & 598 & -214.60   & 20  & 11.33\% \\
\rowcolor{orange!14}
LLama-3.1-Nemotron  & NVIDIA & 70B & llama3.1    &\xmark & 109 & 21   & 620 & -226.20   & 3  & 9.33\% \\

\rowcolor{orange!17}
Codegemma     & Google & 7B & gemma         & \xmark  & 81  & 0    & 669 & -251.4   & 0  & 8\%     \\
\rowcolor{orange!21}
Mistral NeMo   & Mistral AI & 12B & Apache-2.0 & \xmark & 84  & 15   & 651 & -243.6   & 1  & 7.33\% \\
\rowcolor{orange!23}
Gemini-1.5-pro      & Google & N/A & Proprietary       & \xmark & 80  & 0    & 670 & -252.0   & 4  & 6.00\%     \\
\rowcolor{orange!25}
Llama3.1      & Meta & 8B & llama3.1         & \xmark & 65  & 0    & 685 & -261.0   & 0  & 4.67\%  \\
\rowcolor{orange!28}
Gemini-1.5-flash     & Google & N/A & Proprietary        & \xmark & 60  & 0    & 690 & -264.0   & 4  & 4.67\%     \\
\rowcolor{orange!31}
Dolphin-2.8 Mistral    & Cognitive & 7B &  Apache-2.0  & \xmark  & 63  & 22   & 665 & -253.4   & 5  & 4\%     \\

\rowcolor{orange!33}
Mistral-openorca   & Microsoft  & 7B & Apache-2.0   & \xmark & 66  & 0    & 684 & -260.4   & 4  & 4\%     \\
\rowcolor{orange!35}
WizardLM2       & Microsoft & 7B &  Apache-2.0      & \xmark  & 57  & 1    & 692 & -265.4   & 4  & 4\%     \\
\rowcolor{orange!36}
CodeQwen1.5       & Qwen & 7B & tongyi-qianwen
         & \xmark  & 65  & 22   & 663 & -252.2   & 7  & 3.33\%  \\

\rowcolor{orange!37}
Qwen2.5    &   Qwen  & 7B & Apache-2.0            & \xmark  & 62  & 9    & 684 & -261.20   & 3  & 3.33\%       \\

\rowcolor{orange!38}
Mixtral-8x7B   & Mistral AI & 47B & Apache-2.0     & \xmark & 60  & 0    & 690 & -264.0   & 5  & 3.33\%  \\
\rowcolor{orange!41}
Gemma          & Google & 7B & gemma         & \xmark  & 53  & 0    & 697 & -268.2   & 2  & 3.33\%  \\

\rowcolor{orange!48}
llava-v1.5       & liuhaotian & 7B & llama2      & \xmark  & 48  & 0    & 702 & -271.2   & 2  & 2.67\%  \\
\rowcolor{red!30} 
\hline
\rowcolor{red!30}
Phi3        & Microsoft & 3B &  MIT          & \xmark  & 26  & 2    & 722 & -283.6   & 0  & 2\%       \\
\rowcolor{red!30}
Codellama     & Meta & 7B & llama2         & \xmark & 19  & 1    & 730 & -288.2   & 0  & 0\%       \\
\rowcolor{red!30}
Deepseek-coder & Depseek AI & 33B & deepseek         & \xmark  & 19  & 1    & 730 & -288.2   & 0  & 0\%       \\
\rowcolor{red!30}
Qwen2.5    &   Qwen  & 3B & Apache-2.0            & \xmark  & 16  & 9    & 725 & -286.8   & 0  & 0\%       \\
\hline
\rowcolor{red!50}
Orca-mini     & Microsoft & 3B &  llama             & \xmark  & 5   & 0    & 745 & -297.0   & 0  & 0\%       \\
\rowcolor{red!50}
Wizard-Vicuna-Uncensored & Cognitive & 7B &  llama       & \xmark & 0 & 4 & 746 & -298.4   & 0  & 0\%       \\
\rowcolor{red!50}
Llama2-Uncensored   & Cognitive & 7B &  llama2   & \xmark  & 0   & 1    & 749 & -299.6   & 0  & 0\%       \\
\bottomrule
\end{tabular}
\smallskip

\textit{*Tool usage means that ChatGPT-4o can execute codes, access the internet, and run bash commands within a sandboxed environment.\\
$\dagger$ The o1-mini model attempted to skip 52 additional tasks but failed to provide the correct XML format. See Section~\ref{sec:limit} for detailed comments.}

\label{tab:llm_performance}
\end{table*}
We evaluated 25 models, with the complete list available in Table~\ref{tab:llm_performance}. Using \( k=5 \), we generated five different questions per template, resulting in a total of 750 questions. Higher \( k \) values can provide an even more accurate assessment.

Some questions templates produce a particularly large prompt (e.g., containing a Base64-encoded image), therefore certain questions can exceed the token limit of what some models can handle. For example a full prompt from question template $135$ results in $17,667$ tokens~\footnote{\url{https://platform.openai.com/tokenizer}}. 

Among the tested models, currently only ChatGPT-4o has the capability to use tools. It can write and run code, access the web, as well as it has support for multimodal data like visual images, voice, and textual data.
\emph{Notably, existing literature often overlooks whether the LLMs under examination possess tool-usage capabilities, and many benchmarks fail to specify whether the chat or API version of a model is employed.}

This section examines common mistakes made by models and highlights model-specific weaknesses. Particular attention is given to OpenAI's models, as they provide the best ground for comparing a tool enabled model (ChatGPT-4o) and the equivalent API model (GPT-4o) that lacks this functionality. While it might initially appear that ChatGPT-4o's performance advantage stems solely from its ability to use tools, a closer analysis reveals that the performance gap between OpenAI's models is influenced by factors beyond tool availability alone.

\subsection{An Enthusiasm for Mathematics}

One notable observation is the persistent eagerness of most non-tool-using API models to engage with mathematical tasks, despite their consistent inability to solve them correctly, as illustrated in Figure~\ref{fig:performance}. This behavior suggests a lack of meta-cognitive awareness, particularly in acknowledging their limitations with tasks requiring tool usage. Despite prompts encouraging these models to refrain from attempting problems they cannot solve, they rarely do so, resulting in unsuccessful attempts at mathematical problems.

A detailed examination of OpenAI's models reveals that GPT-4o, GPT-4o-mini, and ChatGPT-4o exhibit similar levels of enthusiasm in attempting such tasks, possibly due to their shared underlying architecture. While GPT-4o and GPT-4o-mini consistently fail at mathematical problems due to the lack of code execution ability, ChatGPT-4o demonstrates proficiency in solving these questions.

This observation aligns with findings by researchers at Apple~\cite{b14}, noting that LLMs mostly rely on sophisticated pattern matching rather than genuine logical reasoning. As shown in Figure~\ref{fig:performance}
and Table~\ref{tab:llm_performance}, most models are incapable of skipping tasks, even when these tasks are infeasible to solve without tools. However, the dataset in~\cite{b14} uses \textit{``relatively simple grade-school math questions, requiring only basic arithmetic operations''}, and doesn't allow non-tool using models to skip.

However, using the DIA framework, OpenAI's o1-mini model presents a completely different picture. When given the option to skip questions, o1-mini conducts an assessment and refrains from attempting to answer tasks it deems beyond its reach, demonstrating significant meta-cognitive awareness in identifying which tasks to avoid. It is important to note that in 52 instances, o1-mini attempted to skip a question but failed to answer $\texttt{I-DO-NOT-KNOW}$ in the specified format, resulting in the response being marked as incorrect (this issue is discussed further in Section~\ref{sec:limit}). These instances indicate that o1-mini's capabilities may be more robust than reported. However, it achieves a high NMS, score ($4/5$ on 14 templates), indicating that its reliability is far from perfect.

\subsection{Cybersecurity vs. Mathematics}

The difference in model behavior between mathematical and cybersecurity tasks offers interesting insights. API models without tool-using capabilities, which readily attempt mathematical problems, show more restraint with cybersecurity tasks, as illustrated in Figure~\ref{fig:performance}. Notably, GPT-4o skips questions nearly 20 times more often in the cybersecurity domain (3 vs. 67). Similarly, models like \emph{Mistral-Nemo} or \emph{GPT-4o-mini}, which never skipped mathematical tasks, skipped cybersecurity tasks 15 and 29 times, respectively.
This suggests that these more complex tasks invoke a higher degree of self-reflection and caution. This behavior indicates that some models are capable of critical evaluation, particularly when the task is complex enough to warrant skipping. For certain cybersecurity prompts, some models objected to answering, which category is subject to False Refusal Rate, however measuring this was outside our focus.

\subsection{Limitations in Current Metrics}

In the traditional setting, skipping is an inferior approach to hallucination or guessing. To underline this, we should examine ChatGPT-4o's performance using a traditional metric like Pass@k. On the DIA-Bench, ChatGPT-4o achieved an impressive 73.3\% Pass@5, meaning it was able to provide a correct answer to $73.3\%$ of the question templates when given five attempts. At the same time, it was only able to ace all five instances of $38.67\%$ of the templates. This artificially boosts the success rate, but does not accurately reflect a model’s reliability and confidence. A similar gap exists between most other well performing models, as can be seen in Table~\ref{tab:metrics}.
\begin{table}[h!]
\centering
\begin{tabular}{lcc}
\toprule
\textbf{Model} & \textbf{Pass@5 (\%)} & \textbf{Conf@5 (\%)} \\
\midrule
ChatGPT-4o & 73.3 & 38.67 \\
GPT-4o API & 34.0 & 17.33 \\
o1-mini    & 48.0 & 21.34 \\
\bottomrule
\end{tabular}
\caption{Performance metrics for ChatGPT-4o, GPT-4o API, and o1-mini.}
\label{tab:metrics}
\end{table}

The significant gap between the Pass@k score and the \emph{Confidence Index} highlights the importance of reliability metrics. At the same time, if someone only focuses on the number of correct solutions they would be led to believe that ChatGPT-4o is a superior model to o1-mini. However, this is the result of ChatGPT capitalizing on the mathematics tasks where it was able to use tools. Regardless, o1-mini beats its tool equipped colleague in achieving the best \emph{Reliability Score}.

\subsection{A Lack of Confidence}

ChatGPT-4o recorded a high \emph{Neaer Miss Score}, achieving 4 out of 5 on 14 templates. We anticipate that as $k$ increases, NMS would rise, lowering the confidence index. During early experimentation with prompt templates, we observed that models often approached tasks with varying strategies. For example, when tasked with adding two smaller numbers, ChatGPT-4o might not immediately write and execute Python code—considering the task trivial—but instead produce an answer close to the correct one. For more complex tasks, it demonstrated this behavior less. 
The highest \emph{Neaer Miss Score} belongs to GPT-4o, achieving $4/5$ in 29 templates. This again highlights the risks of traditional metrics like Pass@k.

ChatGPT-4o completely skipped only 9 out of 150 templates, but inconsistently skipped at least one instance in 32 other templates, revealing gaps in its decision-making. This suggests that its self-assessment mechanism isn't fully reliable, as it engages with tasks that later recognizes as too difficult.

In consistent skipping, o1-mini takes the lead with $23$ question templates completely skipped, GPT-4o managed $7$, while GPT-4o-mini was only able to stay consistent for $4$ templates. 
Despite this, ChatGPT-4o outperforms its API counterparts, which is expected since API models lack tools and code execution capabilities. Interestingly, both GPT-4o and ChatGPT-4o exhibit similar ``taste'' when attempting task categories, with GPT-4o taking on math problems enthusiastically despite lacking tools.
In cybersecurity tasks, GPT-4o and ChatGPT-4o display a strong correlation in their skipping patterns—when GPT-4o skips a template, ChatGPT-4o often does the same for at least some instances of that template.

Other API models rarely utilized the skip option and attempted to answer most questions. This pattern, evident in Table~\ref{tab:llm_performance}, underscores that most models are primarily engaged in pattern matching and remain distant from achieving artificial general intelligence (AGI).
The results reveal a substantial gap in problem-solving performance between models with tool-using capabilities and those without. LLMs without tools tend to hallucinate more often, failing to complete tasks accurately, which is reflected in their lower reliability scores.

Most importantly, we observe that the mere availability of tools does not enhance a model's overall judgment. ChatGPT-4o and GPT-4o exhibit similar performance and a comparable tendency in attempting tasks. However, o1-mini, despite lacking tools, surpasses both in reliability and self-reflection.

\section{Limitations and Threats to Validity}
\label{sec:limit}

The accuracy and reliability of benchmarking outcomes depend on the quality of the generator scripts, and despite thorough proof checking on our part, some errors or edge cases may persist. If any issues are found, please report them by opening a GitHub issue.

Certain task templates might not be as well represented as others, leading to an incomplete evaluation of the LLM's overall performance. In our study, we focused on computer science and mathematics, omitting other disciplines to economize our resources. We aim to expand this, and the used data modalities.

We did not utilize o1-preview or other chat models due to their time and cost requirements, which would have impacted the feasibility of this work.
 
During the evaluation, we noticed that several models, particularly o1-mini, attempted to skip tasks but did not follow the required format, instead providing responses like: \emph{\texttt{```xml\textbackslash{}nI-DO-NOT-KNOW\textbackslash{}n```}}. After careful consideration, we decided not to accept such cases, as this presents a limitation in model's capabilities.

\section{Conclusion}
\label{sec:conclusion}

In this study, we introduced the Dynamic Intelligence Assessment (DIA) framework along with the DIA-Bench dataset to test the problem-solving reliability and confidence of Large language Models (LLMs) for mathematics and computer science questions. We also introduced four new key metrics for evaluating LLMs' reliability and confidence. Our work addressed three key research questions:

\begin{itemize} \small 

\item \textbf{RQ1}: \textit{What metrics can best evaluate a model’s confidence and reliability in problem-solving?}
\textbf{Answer}: Traditional metrics often provide a misleading picture of a model’s true performance, as they don't account for consistency. We introduce four metrics: the \textit{Reliability Score}, which penalizes incorrect answers, the \textit{Task Success Rate} (TSR) for measuring consistency across repeated instances, the \textit{Confidence Index} to assess performance across variations of the same task, and the \textit{Near Miss Score} for tasks where the model almost but not fully succeeds. 

\item \textbf{RQ2}: \textit{How can a benchmark dataset better facilitate the accurate measurement of LLMs’ confidence and reliability in problem-solving?}

\textbf{Answer}: By using dynamic question templates, hosting simpler tasks to very difficult ones across various formats (text, PDFs, compiled binaries, etc.). This, combined with the use of the proposed evaluation metrics, offers a more rigorous comparison of models' abilities compared to static, non-multimodal, and overly easy benchmarks, with metrics like Pass@k or Accuracy.

\item \textbf{RQ3}: \textit{How does the use of tools impact the confidence and reliability of various models in problem-solving?}

\textbf{Answer}: 
While the use of tools enhances a model's problem-solving ability, it doesn't necessarily improve its judgment or self-assessment ability. GPT-4o and ChatGPT-4o share the same underlying architecture, and ChatGPT-4o performs better primarily because it has access to tools, not because of better chain-of-thought type reasoning. GPT-4o often attempts tasks it has no chance to solve, showing a lack of judgment. In contrast, o1-mini, a non-tool-using API model, demonstrates superior reasoning by effectively deciding when not to attempt unsolvable tasks, leading to a higher \textit{Reliability Score} than the tool-using ChatGPT-4o. Therefore, while tools improve problem-solving performance, the model's design and architecture are the critical factors determining confidence and reliability.

\end{itemize}

Our findings demonstrate that while current LLMs have made significant progress, there remain challenges to achieve  reliable problem-solving capabilities.
OpenAI's o1 model family is to this date the first among the tested models, that exhibit capabilities beyond just simple pattern matching, and demonstrated excellent critical thinking and self-reflection.

\section*{Acknowledgement}
This research is supported and funded by the Technology Innovation Institute (TII), Abu Dhabi. Additional support is provided by the TKP2021-NVA Funding Scheme under Project TKP2021-NVA-29; ELTE-OTP Cyberlab—a collaboration between Eötvös Loránd University (ELTE) and OTP Bank Plc; EPSRC grant EP/T026995/1 titled “EnnCore: End-to-End Conceptual Guarding of Neural Architectures” under the Security for All in an AI-enabled Society program; the Research Council of Norway Project No. 312122 “Raksha: 5G Security for Critical Communications”; funding from Horizon Europe under Grant Agreement No. 101120853; and funding under Grant Agreement No. 101145874, supported by the European Cybersecurity Competence Centre.


\begin{thebibliography}{00}
\bibitem{b1} A. M. Turing, H. R. Lewis, ``Computing Machinery and Intelligence," The MIT Press, 2021. DOI: 10.7551/mitpress/12274.003.0016.
\bibitem{b2} M. Campbell, A. J. Hoane Jr., and F.-h. Hsu, "Deep blue," Artificial Intelligence, vol. 134, pp. 57--83, 2002.
\bibitem{b3} F.-h. Hsu, ``Behind Deep Blue: Building the Computer That Defeated the World Chess Champion," Princeton University Press, 2022. DOI: 10.2307/j.ctv22pzxz1.
\bibitem{b4} J. J. Hopfield, ``Neural networks and physical systems with emergent collective computational abilities," Proceedings of the National Academy of Sciences, vol. 79, pp. 2554--2558, 1982.
\bibitem{b5} D. E. Rumelhart, G. E. Hinton, and R. J. Williams, ``Learning representations by back-propagating errors,'' Nature, vol. 323, pp. 533--536, 1986.

\bibitem{b6} A. Vaswani, N. Shazeer, N. Parmar, J. Uszkoreit, L. Jones, A. N. Gomez, et al., ``Attention is all you need," in Advances in Neural Information Processing Systems, vol. 30, Curran Associates, Inc., 2017. [Online]. Available: \url{https://proceedings.neurips.cc/paper/2017/file/3f5ee243547dee91fbd053c1c4a845aa-Paper.pdf}.

\bibitem{b7} J. Devlin, M.-W. Chang, K. Lee, and K. Toutanova, ``Bert: Pre-training of deep bidirectional transformers for language understanding," arXiv:1810.04805, 2018.


\bibitem{b8} T. Brown, B. Mann, N. Ryder, M. Subbiah, J. D. Kaplan, P. Dhariwal, et al., ``Language models are few-shot learners," in Advances in Neural Information Processing Systems, vol. 33, Curran Associates, Inc., 2020, pp. 1877--1901.

\bibitem{b9} A. Wang, A. Singh, J. Michael, F. Hill, O. Levy, and S. R. Bowman, ``GLUE: A multi-task benchmark and analysis platform for0 natural language understanding," in Proceedings of the International Conference on Learning Representations (ICLR), 2019.

\bibitem{b10} P. Rajpurkar, J. Zhang, K. Lopyrev, and P. Liang, ``SQuAD: 100,000+ questions for machine comprehension of text," in Proceedings of the 2016 Conference on Empirical Methods in Natural Language Processing, J. Su, K. Duh, et al., Eds., Austin, TX: Association for Computational Linguistics, Nov. 2016, pp. 2383--2392. DOI: 10.18653/v1/D16-1264. [Online]. Available: https://aclanthology.org/D16-1264.

\bibitem{b11} M. Chen, J. Tworek, H. Jun, Q. Yuan, H. P. O. Pinto, J. Kaplan, et al., ``Evaluating large language models trained on code," arXiv preprint arXiv:2107.03374, 2021.

\bibitem{b12} S. Wang, Z. Long, Z. Fan, Z. Wei, and X. Huang, ``Benchmark self-evolving: A multi-agent framework for dynamic LLM evaluation," arXiv:2402.11443, 2024.
\bibitem{b13} S. Honarvar, M. Wilk, and A. Donaldson, ``Turbulence: Systematically and automatically testing instruction-tuned large language models for code," arXiv:2312.14856, 2023.
\bibitem{b14} I. Mirzadeh, K. Alizadeh, H. Shahrokhi, O. Tuzel, S. Bengio, and M. Farajtabar, ``GSM-Symbolic: Understanding the limitations of mathematical reasoning in large language models," arXiv preprint arXiv:2410.05229, 2024.
\bibitem{b15} K. Papineni, S. Roukos, T. Ward, W.-J. Zhu, P. Isabelle, and E. Charniak, ``Bleu: A method for automatic evaluation of machine translation," in Proceedings of the 40th Annual Meeting of the Association for Computational Linguistics, P. Isabelle, Ed., Philadelphia, PA, USA: Association for Computational Linguistics, Jul. 2002, pp. 311--318. DOI: 10.3115/1073083.1073135. [Online]. Available: https://aclanthology.org/P02-1040.
\bibitem{b16} C.-Y. Lin, ``Rouge: A package for automatic evaluation of summaries," in Text summarization branches out, 2004, pp. 74--81.
\bibitem{b17} T. Zhang, V. Kishore, F. Wu, K. Q. Weinberger, and Y. Artzi, ``Bertscore: Evaluating text generation with BERT," arXiv:1904.09675, 2019.
\bibitem{b18} N. Carlini and D. Wagner, ``Towards evaluating the robustness of neural networks," in 2017 IEEE Symposium on Security and Privacy (SP), 2017, pp. 39--57. DOI: 10.1109/SP.2017.49.
\bibitem{b19} J. Liu, Z. Shen, Y. He, X. Zhang, R. Xu, H. Yu, et al., ``Towards out-of-distribution generalization: A survey," arXiv:2108.13624, 2021.
\bibitem{b20} M. Bhatt, S. Chennabasappa, Y. Li, C. Nikolaidis, D. Song, S. Wan, et al., ``Cyberseceval 2: A wide-ranging cybersecurity evaluation suite for large language models," arXiv:2404.13161, 2024.
\bibitem{b21} S. Wang, Z. Li, H. Qian, C. Yang, Z. Wang, M. Shang, et al., ``ReCode: Robustness evaluation of code generation models," in ACL 2023, 2022. [Online]. Available: https://www.amazon.science/publications/recode-robustness-evaluation-of-code-generation-models.
\bibitem{b22} N. Tihanyi, M. A. Ferrag, R. Jain, T. Bisztray, and M. Debbah, ``CyberMetric: A benchmark dataset based on retrieval-augmented generation for evaluating LLMs in cybersecurity knowledge," in 2024 IEEE CSR, 2024, pp. 296--302. DOI: 10.1109/CSR61664.2024.10679494.
\bibitem{b23} H. Yu, B. Shen, D. Ran, J. Zhang, Q. Zhang, Y. Ma, et al., ``Codereval: A benchmark of pragmatic code generation with generative pre-trained models," in 46th IEEE/ACM ICSE, 2024, pp. 1--12.
\bibitem{b24} N. Jiang, T. Lutellier, and L. Tan, ``Cure: Code-aware neural machine translation for automatic program repair," in 2021 IEEE/ACM 43rd ICSE, 2021, pp. 1161--1173.
\bibitem{b25} x.ai, ``Grok-2 beta release," Accessed: 2024-10-06, 2024. [Online]. Available: https://x.ai/blog/grok-2


\bibitem{b26} S. Reddy, D. Chen, and C. D. Manning, ``CoQA: A conversational question answering challenge," Trans. Assoc. Comput. Linguistics, vol. 7, pp. 249-266, 2019.


\bibitem{b27} Z. Yang, P. Qi, S. Zhang, Y. Bengio, W. W. Cohen, R. Salakhutdinov, et al., ``HotpotQA: A dataset for diverse, explainable multi-hop question answering," arXiv:1809.09600, 2018.




\bibitem{b28} X. Wang, Z. Wang, J. Liu, Y. Chen, L. Yuan, H. Peng, et al., ``MINT: Evaluating LLMs in multi-turn interaction with tools and language feedback," arXiv:2309.10691, 2023.

\bibitem{b29} D. Hendrycks, C. Burns, S. Basart, A. Critch, J. Li, D. Song, et al., ``Aligning AI with shared human values," in Proceedings of the International Conference on Learning Representations (ICLR), 2021.

\bibitem{b30} D. Hendrycks, C. Burns, S. Basart, A. Zou, M. Mazeika, D. Song, et al., ``Measuring massive multitask language understanding," in Proceedings of the International Conference on Learning Representations (ICLR), 2021.

\bibitem{b31} Y. Wang, X. Ma, G. Zhang, Y. Ni, A. Chandra, S. Guo, et al., ``Mmlu-pro: A more robust and challenging multi-task language understanding benchmark," arXiv:2406.01574, 2024.

\bibitem{b32} X. Yue, Y. Ni, K. Zhang, T. Zheng, R. Liu, G. Zhang, et al., ``Mmmu: A massive multi-discipline multimodal understanding and reasoning benchmark for expert AGI," in Proceedings of the IEEE/CVF Conference on Computer Vision and Pattern Recognition, 2024, pp. 9556--9567.

\bibitem{b33} M. Mathew, D. Karatzas, and C. V. Jawahar, ``Docvqa: A dataset for VQA on document images," in Proceedings of the IEEE/CVF Winter Conference on Applications of Computer Vision, 2021, pp. 2200--2209.

\bibitem{b34} J. Austin, A. Odena, M. Nye, M. Bosma, H. Michalewski, D. Dohan, et al., ``Program synthesis with large language models," arXiv:2108.07732, 2021.

\bibitem{b35} D. Hendrycks, C. Burns, S. Kadavath, A. Arora, S. Basart, E. Tang, et al., ``Measuring mathematical problem solving with the MATH dataset," in Advances in Neural Information Processing Systems (NeurIPS), 2021.

\bibitem{b36} P. Lu, H. Bansal, T. Xia, J. Liu, C. Li, H. Hajishirzi, et al., ``MathVista: Evaluating mathematical reasoning of foundation models in visual contexts," in Proceedings of the International Conference on Learning Representations (ICLR), 2024.

\bibitem{b37} D. Hendrycks, S. Basart, S. Kadavath, M. Mazeika, A. Arora, E. Guo, et al., ``Measuring coding challenge competence with APPS," in Advances in Neural Information Processing Systems (NeurIPS), 2021.

\bibitem{b38} M. Kaufmann, D. Kang, Y. Sun, S. Basart, X. Yin, M. Mazeika, et al., ``Testing robustness against unforeseen adversaries," arXiv:1908.08016, 2019.

\bibitem{b39} A. Athalye, L. Engstrom, A. Ilyas, and K. Kwok, ``Synthesizing robust adversarial examples," in Proceedings of the International Conference on Machine Learning, 2018, pp. 284--293.

\bibitem{b40} J.-B. Döderlein, M. Acher, D. E. Khelladi, and B. Combemale, ``Piloting copilot and codex: Hot temperature, cold prompts, or black magic?," arXiv preprint arXiv:2210.14699, 2022.

\bibitem{b41} Z. Liu, ``SecQA: A Concise Question-Answering Dataset for Evaluating Large Language Models in Computer Security," arXiv preprint arXiv:2312.15838, 2023.


\bibitem{b42} H. Pearce, B. Ahmad, B. Tan, B. Dolan-Gavitt, and R. Karri, ``Asleep at the keyboard? Assessing the security of GitHub Copilot’s code contributions," in 2022 IEEE Symposium on Security and Privacy (SP), 2022, pp. 754--768.

\bibitem{b43} N. Tihanyi, T. Bisztray, R. Jain, M. A. Ferrag, L. C. Cordeiro, and V. Mavroeidis, ``The FormAI dataset: Generative AI in software security through the lens of formal verification," in Proceedings of the 19th International Conference on Predictive Models and Data Analytics in Software Engineering, 2023, pp. 33--43.
\newpage
\bibitem{b44} K. Cobbe, V. Kosaraju, M. Bavarian, M. Chen, H. Jun, L. Kaiser, et al., ``Training verifiers to solve math word problems," arXiv preprint arXiv:2110.14168, 2021.


\bibitem{b45} W.-C. Kwan, X. Zeng, Y. Wang, Y. Sun, L. Li, L. Shang, et al., ``M4LE: A multi-ability multi-range multi-task multi-domain long-context evaluation benchmark for large language models," arXiv preprint arXiv:2310.19240, 2023.

\bibitem{b46} Z. Dong, T. Tang, J. Li, W. X. Zhao, and J.-R. Wen, ``Bamboo: A comprehensive benchmark for evaluating long text modeling capacities of large language models," arXiv preprint arXiv:2309.13345, 2023.

\bibitem{b47} Y. Bai, X. Lv, J. Zhang, H. Lyu, J. Tang, Z. Huang, et al., ``LongBench: A bilingual, multitask benchmark for long context understanding," arXiv preprint arXiv:2308.14508, 2023.

\bibitem{b48} D. Song, S. Chen, G. H. Chen, F. Yu, X. Wan, and B. Wang, ``Milebench: Benchmarking MLLMs in long context," arXiv preprint arXiv:2404.18532, 2024.




\end{thebibliography}
\end{document}